\documentclass[letterpaper]{article} 
\usepackage{aaai2027}  
\usepackage[hyphens]{url}  
\usepackage{graphicx} 
\usepackage{natbib}  
\usepackage{caption} 

\usepackage{algorithm}
\usepackage{algpseudocode}

\algrenewcommand\alglinenumber[1]{\scriptsize #1:}
\usepackage{amsmath}
\usepackage{amssymb}
\usepackage{booktabs}
\usepackage{tabularx}
\usepackage{dblfloatfix}
\usepackage[most]{tcolorbox}
\definecolor{figink}{HTML}{1F2329}
\definecolor{figaccent}{HTML}{3347E6}
\definecolor{figaccentpale}{HTML}{EAEDFD}
\tcbset{promptbox/.style={
    enhanced,
    colback=figaccentpale!45!white,
    colframe=figaccent,
    colbacktitle=figaccent,
    coltitle=white,
    colupper=figink,
    fonttitle=\bfseries\footnotesize,
    fontupper=\footnotesize,
    boxrule=0.7pt,
    arc=1.5mm,
    left=2mm,
    right=2mm,
    top=1.3mm,
    bottom=1.3mm}}
\usepackage{newfloat}
\usepackage{listings}
\DeclareCaptionStyle{ruled}{labelfont=normalfont,labelsep=colon,strut=off} 
\floatstyle{ruled}
\newfloat{listing}{tb}{lst}{}
\floatname{listing}{Listing}

\title{CSPO: Hidden-State Cloud Geometry for \\Semi-Supervised Preference Optimization in LLM Reasoning}
\author{
    Yuzhou Liu\textsuperscript{\rm 1},
    Xiyang Hu\textsuperscript{\rm 2}
}
\affiliations{
    \textsuperscript{\rm 1}University of Southern California\\
    \textsuperscript{\rm 2}Arizona State University\\
    yuzhoul@usc.edu, xiyanghu@asu.edu
}

\begin{document}
\maketitle

\begin{abstract}
Preference optimization improves mathematical reasoning in large language models (LLMs), yet building reliable preference pairs typically requires costly external supervision. We investigate whether preference supervision can instead be derived from a model's internal representation geometry in a semi-supervised setting. Across distinct math problems, reasoning trajectories form global point clouds in which correct solutions cluster more densely and merge into connected components at smaller distance scales than incorrect ones. Leveraging this structure, we propose Cloud-guided Semi-supervised Preference Optimization (CSPO). From a small labeled set, CSPO embeds verified correct and incorrect rollouts by the normalized mean of their final-layer response-token hidden states and samples multiple pairs of correct and incorrect reference clouds. To mine preference pairs on a much larger unlabeled set, CSPO samples multiple rollouts per unlabeled problem and scores each one against these labeled reference clouds, contrasting its soft $k$-nearest-neighbor proximity to early connected components in the correct versus incorrect clouds. For each unlabeled problem, majority voting over sampled answers selects the preferred and rejected answers, while the cloud score chooses the highest-scoring majority trajectory and lowest-scoring minority trajectory and retains only pairs with the largest score gap. Across four model settings on GSM8K and MATH-Numeric, CSPO outperforms strong semi-supervised baselines by up to 4.49 and 4.19 percentage points, respectively, using far fewer pairs than an external reward model. Pair-level diagnostics confirm that CSPO preserves correctness reliability while more cleanly separating well-formed chosen trajectories from incomplete, repetitive, or corrupted rejections.
\end{abstract}


\section{Introduction}
Post-training and preference optimization substantially strengthen mathematical and symbolic reasoning in large language models (LLMs), with prominent methods including reinforcement learning from human feedback (RLHF) \citep{ouyang2022training}, direct preference optimization (DPO) \citep{rafailov2023direct}, and group relative policy optimization (GRPO) \citep{shao2024deepseekmath}. The quality of the resulting policy depends heavily on how preferred and rejected responses are constructed \citep{christiano2017deep, stiennon2020learning, ouyang2022training}. Human preference labels are costly, external verifiers are often unavailable, and final-answer supervision alone reveals little about the quality of intermediate reasoning steps, especially when each prompt yields multiple long rollouts.

In semi-supervised reasoning, only a small subset of prompts has verified answers while most prompts and generated trajectories remain unlabeled. Prior approaches lower annotation costs by learning from model-generated reasoning: STaR iteratively fine-tunes on self-generated rationales that reach verified answers \citep{zelikman2022star}, self-consistency selects the most frequent final answer across multiple sampled rollouts via majority voting at inference time \citep{wang2023selfconsistencyimproveschainthought}, and ScPO turns self-consistency into training supervision by preferring trajectories from the majority-answer cluster over minority-answer alternatives \citep{prasad2025selfconsistencypreferenceoptimization}. However, answer-level voting treats all trajectories sharing the same final answer as equivalent, ignoring differences in reasoning quality, and becomes unreliable when a prompt's majority consensus is weak or incorrect. Semi-supervised reward modeling offers an alternative by iteratively training a reward model to pseudo-label unlabeled responses \citep{he2024semisupervised}, yet its supervision quality remains tied to a separately trained reward model and can be brittle under scarce labeled preferences.

We explore a complementary source of preference supervision: the global geometry of the language model's own hidden representations. Reasoning trajectories generated across many distinct mathematical problems do not scatter independently in activation space; instead, they form a structured global point cloud whose organization correlates with solution quality. Correct trajectories from different problems occupy denser, more coherent neighborhoods and merge into connected clusters at smaller distance thresholds (filtration scales), whereas incorrect trajectories are more dispersed and structurally heterogeneous. To capture the full reasoning trace instead of only the terminal token, we represent each rollout by the $\ell_2$-normalized mean of its response-token hidden states.

Grounded in this geometric structure, we introduce Cloud-guided Semi-supervised Preference Optimization (CSPO), a topology-guided preference-mining framework for semi-supervised reasoning optimization. Using a small labeled set, CSPO samples multiple balanced pairs of correct and incorrect reference clouds (reference banks), extracts early connected components from each cloud, and scores every unlabeled rollout via a component-level soft $k$-nearest-neighbor proximity measure averaged across banks. Integrating this global cloud score with per-problem self-consistency enables fine-grained trajectory selection within answer clusters and confidence-based pair filtering without external reward models or gold labels on unlabeled prompts.

Evaluations on GSM8K and MATH-Numeric across multiple model backbones demonstrate that CSPO consistently surpasses supervised fine-tuning (SFT) and ScPO, and achieves competitive or stronger performance than a reward-model baseline using far fewer preference pairs. Diagnostic analyses further reveal that cloud scoring sharpens the contrast between preferred and rejected responses, directing truncated, repetitive, or corrupted rollouts to the rejected side without increasing preference reversals. In summary, our contributions are: (1) identifying a cross-problem geometric signal in trajectory-level hidden states that correlates with reasoning quality; (2) designing a multi-bank, component-level cloud-scoring mechanism grounded in early connectivity; and (3) combining cloud scoring with self-consistency to mine high-confidence preference pairs for semi-supervised reasoning optimization.

\section{Related Work}
\noindent\textbf{Latent-space supervision for reasoning optimization.} Internal representations of language models encode signals linked to latent knowledge, truthfulness, and hallucination risk \citep{burns2024discoveringlatentknowledgelanguage, azaria-mitchell-2023-internal, chen2024insidellmsinternalstates}. Leveraging these signals for reasoning, \emph{Silence the Judge} introduces Latent-GRPO, which derives intrinsic rewards from the hidden-state geometry of rollouts sampled for the same prompt \citep{zhang2026silence}. It estimates a prompt-specific latent center and rewards trajectories closer to that center. While this removes the need for external verifiers, the resulting signal is confined to a single prompt's rollout group and remains vulnerable to incorrect within-prompt consensus. By contrast, CSPO constructs multi-bank reference clouds from labeled trajectories across many problems, transferring global geometric supervision across prompts and combining it with self-consistency to mine high-confidence preference pairs.

\noindent\textbf{Preference optimization from sampled reasoning trajectories.} A growing line of work improves reasoning through self-generated rationales, iterative self-training, and preference-based learning \citep{zelikman2022star,hosseini2024vstartrainingverifiersselftaught,wang2024selftrainingdirectpreferenceoptimization}. In particular, IRPO and ScPO convert multiple sampled chain-of-thought rollouts into pairwise preference supervision. IRPO iteratively pairs trajectories reaching correct answers against incorrect ones and optimizes a DPO objective augmented with a negative log-likelihood term \citep{rafailov2023direct, pang2024iterativereasoningpreferenceoptimization}. ScPO reduces reliance on gold answers by leveraging self-consistency \citep{wang2023selfconsistencyimproveschainthought}, preferring trajectories from the majority-answer cluster over inconsistent alternatives on unlabeled prompts \citep{prasad2025selfconsistencypreferenceoptimization}. However, both methods rely solely on within-prompt outcomes and ignore geometric regularities shared across problems. CSPO retains ScPO's answer-grouping mechanism while adding cross-problem cloud scoring to rank trajectories within answer clusters and filter pairs by their geometric margin.

\noindent\textbf{Topology and geometry of LLM representation spaces.} Recent studies have shown that language-model activations exhibit rich geometric and topological organization. Persistent homology has been used to analyze hidden-state point clouds and connect their structure to model behavior and robustness \citep{chauhan2022bertopsstudyingbertrepresentations}, while linear and geometric analyses have revealed activation-space directions tracking factual truth \citep{marks2024geometrytruthemergentlinear}. More recently, \emph{The Shape of Adversarial Influence} demonstrated that adversarial conditions systematically alter the global topology of LLM activation spaces \citep{fay2026shape}. Whereas these works are primarily diagnostic, we operationalize cross-problem representation geometry into multi-bank reference clouds, converting early connectivity structure into trajectory scores and filtered preference pairs for semi-supervised reasoning optimization.

\section{Method}
\begin{figure*}[!t]
    \centering
    \includegraphics[width=\textwidth]{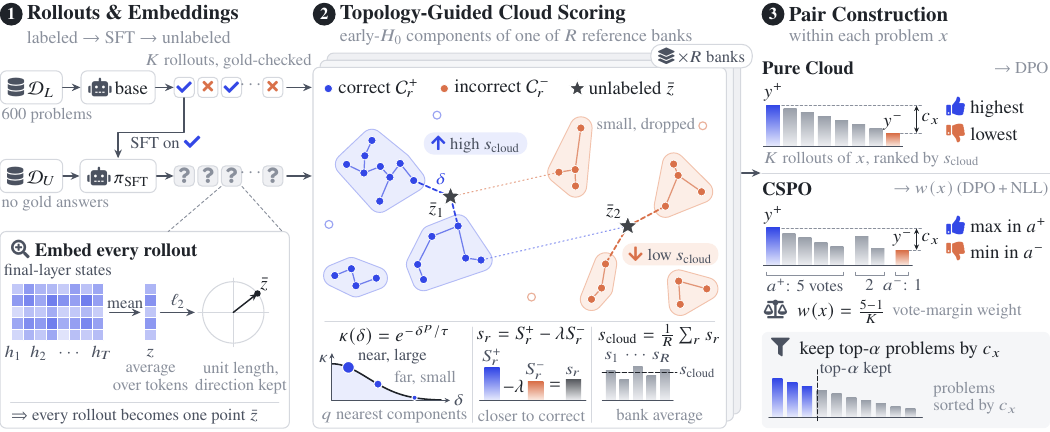}
\caption{Overview of CSPO. \textbf{(1) Rollouts and embeddings:} the base model generates $K$ rollouts for each labeled problem in $\mathcal{D}_L$; rollouts are checked against the gold answers, and the verified correct ones are used for supervised fine-tuning. The tuned policy $\pi_{\mathrm{SFT}}$ then generates $K$ rollouts for each unlabeled problem in $\mathcal{D}_U$. Every rollout is embedded as the $\ell_2$-normalized mean of its final-layer response-token hidden states. \textbf{(2) Topology-guided cloud scoring:} labeled embeddings form correct and incorrect pools, from which $R$ balanced reference banks are sampled; within each cloud, early-$H_0$ merges yield connected components, and small components are discarded. An unlabeled rollout $\bar z$ is scored by a distance kernel over its $q$ nearest components, weighted by component size (only the nearest opposite-cloud component is drawn), contrasted between the correct and incorrect clouds, and averaged over banks into $s_{\mathrm{cloud}}$. \textbf{(3) Pair construction:} within each problem, Pure Cloud pairs the highest- and lowest scoring rollouts and trains with DPO; CSPO pairs the highest-scoring rollout of the majority answer $a^{+}$ with the lowest-scoring rollout of the least-frequent answer $a^{-}$ and trains with vote-weighted DPO\,+\,NLL. Both keep the top-$\alpha$ problems ranked by the cloud margin $c_x$.}
    \label{fig:cloud_pipeline}
\end{figure*}

As illustrated in Figure~\ref{fig:cloud_pipeline}, both CSPO and Pure Cloud follow a semi-supervised pipeline that begins with a base model and a small labeled dataset.
\subsection{Problem Setup}
We consider a semi-supervised reasoning setting with a small labeled set $\mathcal{D}_L=\{(x_i,a_i)\}_{i=1}^{n_L}$ and a larger unlabeled set $\mathcal{D}_U=\{x_j\}_{j=1}^{n_U}$, where $n_L \ll n_U$. We first use the base model to generate $K$ reasoning trajectories for each labeled problem in $\mathcal{D}_L$. Comparing each extracted answer with the gold answer partitions the labeled trajectories into correct and incorrect pools. The verified correct trajectories are used for supervised fine-tuning, producing a tuned policy $\pi_{\mathrm{SFT}}$, while both labeled pools are retained to construct the multi-bank reference clouds. Finally, $\pi_{\mathrm{SFT}}$ generates $K$ trajectories for each problem in $\mathcal{D}_U$. These unlabeled trajectories are scored against the reference clouds and paired into preference data without accessing any ground-truth answers from $\mathcal{D}_U$.

To characterize local connectivity within each reference cloud, we use zero-dimensional persistent homology ($H_0$) under a Vietoris--Rips filtration, which connects pairs of points within an increasing distance threshold. Initially, each trajectory forms its own connected component; as the threshold grows, nearby components merge, and $H_0$ records the scales at which these merges occur \citep{zomorodian2005computing,gabrielsson2020topology}. Persistent-homology visualizations, including exploratory $H_1$ (loop-structure) analyses and the empirical motivation for our component construction, appear in Appendix~A.

\subsection{Trajectory Representations and Reference Clouds}

For a response trajectory $y$ to problem $x$ consisting of $T$ valid response tokens, let $h_t\in\mathbb{R}^d$ be the final-layer hidden state of the $t$-th token, where $d$ is the hidden dimension. We represent the trajectory by its $\ell_2$-normalized mean hidden state:
\begin{equation}
\bar z(x,y)
=
\frac{z(x,y)}{\|z(x,y)\|_2},
\qquad
z(x,y)=\frac{1}{T}\sum_{t=1}^{T}h_t .
\label{eq:trajectory-representation}
\end{equation}
Equation~(\ref{eq:trajectory-representation}) summarizes the full reasoning trace instead of relying solely on the terminal token.

Using the labeled trajectories, we form a correct pool $\mathcal{P}^{+}$ and an incorrect pool $\mathcal{P}^{-}$ according to whether their extracted answers match the verified answers. We then construct $R$ balanced reference banks $(\mathcal{C}_r^{+},\mathcal{C}_r^{-})$ by repeatedly sampling from these two pools. This cross-problem construction contrasts with prompt-specific clouds formed from multiple responses to a single problem \citep{zhang2026silence}.

\subsection{Topology-Guided Cloud Scoring}

Each reference cloud may contain multiple local clusters. To identify them, we process pairwise edges in ascending Euclidean distance, as in the zero-dimensional Vietoris--Rips filtration \citep{gabrielsson2020topology}. Starting with one component per point in a reference cloud of $N=|\mathcal{C}_r^c|$ trajectories, we stop after $\lceil\rho(N-1)\rceil$ successful merges, where $\rho\in(0,1)$ is the merge ratio, and discard components smaller than a predefined threshold. The remaining early-$H_0$ components of class $c\in\{+,-\}$ in bank $r$ are denoted by $\mathcal{G}_r^c=\{G_{r,j}^c\}_j$.

For a candidate $z$, let $\delta_{r,j}^c(z)=\min_{u\in G_{r,j}^c}\|z-u\|_2$ be its Euclidean distance to component $G_{r,j}^c$. Let $\mathcal{N}_{q,r}^c(z)$ contain its $q$ nearest components, and define the component-size weight as
\[
\omega_{r,j}^c = |G_{r,j}^c|,
\]
where $|G_{r,j}^c|$ is the number of reference trajectories in the component. We use the distance kernel
\[
\kappa(\delta)=\exp\left(-\frac{\delta^p}{\tau}\right),
\]
where $\delta\ge 0$ is a distance, $p>0$ controls the distance-decay rate ($p=2$ in all experiments), and $\tau>0$ is the kernel bandwidth parameter. The component compatibility is
\begin{equation}
S_r^c(z)
=
\log
\frac{
 \sum\limits_{j\in\mathcal{N}_{q,r}^c(z)}
 \omega_{r,j}^c\kappa(\delta_{r,j}^c(z))
}{
 \sum\limits_{j\in\mathcal{N}_{q,r}^c(z)}
 \omega_{r,j}^c
}.
\label{eq:component-score}
\end{equation}
Equation~(\ref{eq:component-score}) thus aggregates compatibility over topology-induced components instead of isolated reference points.

The final cloud score contrasts compatibility with the correct and incorrect reference clouds and averages across all $R$ banks:
\begin{equation}
\begin{array}{rcl}
s_{\mathrm{cloud}}(x,y)
&=&\displaystyle\frac{1}{R}
\sum_{r=1}^{R}s_r(\bar z(x,y)),\\[3pt]
s_r(z)
&=&S_r^{+}(z)-\lambda_{\mathrm{neg}}S_r^{-}(z).
\end{array}
\label{eq:cloud-score}
\end{equation}
In Eq.~(\ref{eq:cloud-score}), $\lambda_{\mathrm{neg}}$ controls the incorrect-cloud penalty. Cloud scores are compared only among trajectories generated for the same problem.

\subsection{Preference Pair Construction}

\paragraph{Pure Cloud.}
Let $\mathcal{Y}_x$ be the set of $K$ sampled trajectories for an unlabeled problem $x$, and let $\mathcal{Y}_x^{\mathrm{valid}}\subseteq\mathcal{Y}_x$ denote the trajectories that pass response-validity checks. For each unlabeled problem, Pure Cloud pairs the highest- and lowest-scoring valid trajectories:
\begin{equation}
\begin{array}{rcl}
y_x^{+}
&=&\displaystyle\arg\max_{y\in\mathcal{Y}_x^{\mathrm{valid}}}
s_{\mathrm{cloud}}(x,y),\\
y_x^{-}
&=&\displaystyle\arg\min_{y\in\mathcal{Y}_x^{\mathrm{valid}}}
s_{\mathrm{cloud}}(x,y).
\end{array}
\label{eq:pure-cloud-pair}
\end{equation}
Pairs are ranked by their cloud margin $s_{\mathrm{cloud}}(x,y_x^{+})- s_{\mathrm{cloud}}(x,y_x^{-})$, and only the top-$\alpha$ highest-confidence fraction is retained. Equation~(\ref{eq:pure-cloud-pair}) thus constructs preference pairs without answer-level self-consistency.

\paragraph{CSPO Hybrid.}
We first extract and canonicalize the final answer of each valid trajectory and group trajectories sharing the same answer into answer clusters. Following ScPO, we retain only problems with a unique majority-answer cluster. The majority answer becomes the preferred answer $a_x^{+}$, while a least-frequent eligible non-majority answer becomes the rejected answer $a_x^{-}$. When multiple minority clusters tie for the minimum vote count, we select the cluster whose lowest-scoring trajectory has the smallest cloud score.

Once the preferred and rejected answer clusters are identified, cloud scores select the concrete trajectories within each cluster:
\begin{equation}
y_x^{+}
=
\arg\max_{y\in\mathcal{S}_x^{+}}
s_{\mathrm{cloud}}(x,y),
\quad
y_x^{-}
=
\arg\min_{y\in\mathcal{S}_x^{-}}
s_{\mathrm{cloud}}(x,y),
\label{eq:hybrid-trajectory-selection}
\end{equation}
where $\mathcal{S}_x^{+}$ and $\mathcal{S}_x^{-}$ denote the preferred and rejected answer clusters, respectively. We define the cloud-based confidence margin of the pair as
\begin{equation}
c_x^{\mathrm{Hybrid}}
=
s_{\mathrm{cloud}}(x,y_x^{+})
-
s_{\mathrm{cloud}}(x,y_x^{-}).
\label{eq:hybrid-confidence}
\end{equation}
Eligible pairs are ranked by this margin, and the top-$\alpha$ fraction is retained, where $\alpha$ is a tunable retention hyperparameter. In this way, self-consistency determines the answer-level preference direction, while cloud scoring selects concrete trajectories and filters pairs by geometric separation. Because ScPO and CSPO can select different trajectories prior to final validity filtering, their final pair counts may differ slightly (see Appendix~B).

For optimization, we retain the normalized vote-margin weight used by ScPO:
\begin{equation}
w(x)
=
\frac{
V_x(a_x^{+})-V_x(a_x^{-})
}{K},
\label{eq:vote-weight}
\end{equation}
where $V_x(a)$ is the number of sampled trajectories producing answer $a$, and $K$ is the total number of rollouts for problem $x$. Because tied-majority problems are excluded, every retained pair has a positive vote margin.

Complete pseudocode and implementation details for Pure Cloud and CSPO preference-pair construction are provided in Appendix~B.

\subsection{Preference Optimization}

\paragraph{Pure Cloud optimization.}
Pure Cloud uses standard DPO \citep{rafailov2023direct}. Let $\pi_\theta$ be the policy being optimized and $\pi_{\mathrm{ref}}=\pi_{\mathrm{SFT}}$ be the frozen reference policy. Define
\[
\psi_\theta(x,y)
=
\log\pi_\theta(y\mid x)
-
\log\pi_{\mathrm{ref}}(y\mid x)
\]
and
\[
\Delta \psi_\theta
=
\psi_\theta(x,y^{+})-\psi_\theta(x,y^{-}).
\]
Its objective over the constructed preference dataset $\mathcal{D}_{\mathrm{pref}}$ is
\begin{equation}
\mathcal{L}_{\mathrm{PureCloud}}
=
-\mathbb{E}_{(x,y^{+},y^{-})\sim\mathcal{D}_{\mathrm{pref}}}
\left[
\log\sigma\!\left(\beta\Delta \psi_\theta\right)
\right],
\label{eq:pure-cloud-loss}
\end{equation}
where $\sigma(\cdot)$ is the logistic sigmoid function and $\beta>0$ is the DPO regularization parameter. Equation~(\ref{eq:pure-cloud-loss}) increases the relative likelihood of the cloud-preferred trajectory with respect to the reference policy.

\paragraph{CSPO optimization.}
Following ScPO, CSPO additionally uses a length-normalized negative log-likelihood objective for the chosen response:
\begin{equation}
\ell_{\mathrm{NLL}}(x,y^{+})
=
-\frac{1}{|y^{+}|}
\sum_{t=1}^{|y^{+}|}
\log\pi_\theta
\left(
y_t^{+}\mid x,y_{<t}^{+}
\right),
\label{eq:nll-loss}
\end{equation}
where $|y^{+}|$ is the token length of $y^{+}$, $y_t^{+}$ is its $t$-th token, and $y_{<t}^{+}$ denotes the preceding response tokens. Let
\begin{equation}
\ell_{\mathrm{DPO}}(x,y^{+},y^{-})
=
-\log\sigma\!\left(\beta\Delta \psi_\theta\right).
\label{eq:dpo-loss}
\end{equation}
The final CSPO objective is
\begin{equation}
\mathcal{L}_{\mathrm{CSPO}}
=
\mathbb{E}_{(x,y^{+},y^{-})\sim\mathcal{D}_{\mathrm{pref}}}
\left[
w(x)
\left(
\ell_{\mathrm{DPO}}
+
\lambda_{\mathrm{NLL}}
\ell_{\mathrm{NLL}}
\right)
\right].
\label{eq:hybrid-loss}
\end{equation}

In Eq.~(\ref{eq:hybrid-loss}), the normalized vote margin defined in Eq.~(\ref{eq:vote-weight}) weights both optimization terms, while $\lambda_{\mathrm{NLL}}$ controls the contribution of the chosen-response likelihood objective. The DPO term learns the pairwise preference, whereas the NLL term directly reinforces the cloud-selected trajectory from the majority-answer cluster.

\section{Experiments}
\subsection{Experimental Setup}

\paragraph{Datasets and Metrics.}
We evaluate CSPO on GSM8K and MATH-Numeric. We additionally analyze the cloud-score gap and the reference-bank configuration on GSM8K, and ablate the Pure Cloud trajectory representation on MATH-Numeric.

\begin{itemize}
    \item \textbf{GSM8K.} GSM8K \citep{cobbe2021training} contains approximately 7.5K training problems and 1.3K test problems covering grade-school mathematical reasoning. Following ScPO, we reserve 10\% of the training set as a development set for hyperparameter tuning and checkpoint selection, yielding roughly 6.7K training, 0.8K development, and 1.3K test problems. We report exact-match accuracy of the extracted final numeric answer on the test set.

    \item \textbf{MATH-Numeric.} MATH \citep{hendrycks2021measuring} consists of challenging high-school mathematics competition problems. We retain examples whose final answers can be evaluated via numeric extraction and normalization. Using the same protocol as GSM8K, we reserve 10\% of the filtered training set as a development set and evaluate final exact-match accuracy on the official test split.
\end{itemize}

\paragraph{Base Models.}
On GSM8K, we evaluate Llama-3-8B Base \citep{grattafiori2024llama} and Mistral-7B-v0.3 \citep{jiang2023mistral}. On MATH-Numeric, we evaluate Llama-3-8B Base and Qwen3-8B \citep{yang2025qwen3}. We additionally evaluate Qwen3-4B-Instruct-2507 \citep{yang2025qwen3}, a stronger instruction-tuned reasoning model, in our Pure Cloud ablation to test whether cloud-based signals grow more informative as reasoning capability improves.

\paragraph{Baselines.}
All methods are compared in a semi-supervised setting with $n_L=600$ randomly sampled labeled training problems and the remaining training problems treated as unlabeled.

\begin{itemize}
    \item \textbf{Base Model (Zero-shot CoT).} We evaluate the base model with zero-shot chain-of-thought prompting \citep{kojima2023largelanguagemodelszeroshot} and greedy decoding, without task-specific training.

    \item \textbf{Supervised Fine-Tuning (SFT-600).} The base model generates rollouts for the 600 labeled problems, and trajectories matching the gold answers are used for supervised fine-tuning. This baseline measures the gain from gold-verified supervision on the labeled subset alone.

    \item \textbf{Self-Consistency Preference Optimization (ScPO).} Following \citet{prasad2025selfconsistencypreferenceoptimization}, we group sampled responses by extracted final answer, pair responses from the most and least frequent answer clusters as chosen and rejected, and weight each pair by the normalized frequency gap.

    \item \textbf{Reward-Model-Based Preference Optimization (IRPO-RM).} We implement a semi-supervised adaptation of the reward-model baseline from \citet{prasad2025selfconsistencypreferenceoptimization}. On the 600 labeled problems, gold-answer correctness partitions rollouts into preferred and rejected pools, and ArmoRM-Llama3-8B \citep{wang2024interpretablepreferencesmultiobjectivereward} selects the highest-scoring correct trajectory as chosen and the lowest-scoring incorrect trajectory as rejected. On unlabeled problems, ArmoRM-Llama3-8B scores all sampled responses and pairs the highest- and lowest-scoring trajectories. Each eligible problem contributes at most one pair ranked by its reward-score margin
    \[
    c_x^{\mathrm{IRPO\text{-}RM}}
    =
    s_{\mathrm{RM}}(x,y_x^{+})
    -
    s_{\mathrm{RM}}(x,y_x^{-}),
    \]
    where $s_{\mathrm{RM}}(x,y)$ is the ArmoRM-Llama3-8B reward score, with uniform training weight $1$ for all retained pairs.
\end{itemize}

\paragraph{Hyperparameters.}
For ScPO and CSPO, we sample $K=8$ rollouts per problem, retain the top $\alpha=0.30$ candidate pairs, and set the merge ratio to $\rho=0.2$; Pure Cloud uses $K=16$ and $\alpha=0.10$. Rollouts are sampled with temperature $1.0$, top-$p=0.95$, and dataset-specific maximum lengths. Cloud scoring uses $R=20$ banks with 200 labeled problems per bank, $q=5$, $\tau=2.0$, and $\lambda_{\mathrm{neg}}=1.0$. Unless noted otherwise, DPO training uses $\beta=0.10$, learning rate $5\times10^{-6}$, effective batch size 16, up to 20 epochs, and early-stopping patience 5; CSPO sets $\lambda_{\mathrm{NLL}}=1.0$. All test accuracies use greedy decoding and exact-match evaluation.

\paragraph{Computing Infrastructure.}
All experiments run on four NVIDIA A40 GPUs (48\,GB memory each).
\subsection{Main Results and Analysis}
\subsubsection{Results on GSM8K.}
CSPO achieves the highest accuracy on both GSM8K backbones, outperforming ScPO and IRPO-RM with far fewer preference pairs than reward-model selection.

As reported in Table~\ref{tab:gsm8k_results}, SFT-600 improves over the base model on both backbones, while preference optimization delivers substantially larger gains. ScPO raises accuracy from 41.62\% to 49.74\% on Llama-3-8B and from 12.28\% to 28.43\% on Mistral-7B. CSPO pushes performance further to 52.24\% on Llama-3-8B and 32.92\% on Mistral-7B, improving over ScPO by 2.50 and 4.49 percentage points, respectively.

\paragraph{Pair quality outweighs pair volume.}
Although IRPO-RM trains on substantially more pairs (3,794 on Llama-3-8B and 2,511 on Mistral-7B), it reaches only 46.40\% and 15.23\%, trailing both ScPO-based methods. Combining per-problem self-consistency with cloud-based trajectory scoring thus yields stronger supervision than volume-heavy reward-model selection.

\begin{table}[!htb]
\centering
\caption{Single-rollout test accuracy and number of preference pairs on GSM8K. The best accuracy for each backbone is shown in bold.}
\label{tab:gsm8k_results}
\small
\setlength{\tabcolsep}{3pt}
\renewcommand{\arraystretch}{1.10}
\begin{tabular*}{\columnwidth}{@{\extracolsep{\fill}}lcccc@{}}
\toprule
& \multicolumn{2}{c}{\textbf{Llama-3-8B}}
& \multicolumn{2}{c}{\textbf{Mistral-7B}} \\
\cmidrule(lr){2-3}\cmidrule(lr){4-5}
\textbf{Method}
& \textbf{Acc. (\%)}
& \textbf{\# Pairs}
& \textbf{Acc. (\%)}
& \textbf{\# Pairs} \\
\midrule

Base Model
& 39.50
& --
& 10.31
& -- \\

SFT-600
& 41.62
& --
& 12.28
& -- \\

\midrule

ScPO
& 49.74
& 1,311
& 28.43
& 1,113 \\

IRPO-RM
& 46.40
& 3,794
& 15.23
& 2,511 \\
CSPO
& \textbf{52.24}
& 1,329
& \textbf{32.92}
& 1,163 \\
\bottomrule
\end{tabular*}
\end{table}

\begin{table}[!htb]
\centering
\caption{Single-rollout test accuracy and number of preference pairs on MATH-Numeric. The best accuracy for each backbone is shown in bold, and the second best is underlined.}
\label{tab:math_results}
\small
\setlength{\tabcolsep}{3pt}
\renewcommand{\arraystretch}{1.10}
\begin{tabular*}{\columnwidth}{@{\extracolsep{\fill}}lcccc@{}}
\toprule
& \multicolumn{2}{c}{\textbf{Llama-3-8B}}
& \multicolumn{2}{c}{\textbf{Qwen3-8B}} \\
\cmidrule(lr){2-3}\cmidrule(lr){4-5}
\textbf{Method}
& \textbf{Acc. (\%)}
& \textbf{\# Pairs}
& \textbf{Acc. (\%)}
& \textbf{\# Pairs} \\
\midrule

Base Model
& 8.03
& --
& 53.61
& -- \\

SFT-600
& 21.01
& --
& 55.52
& -- \\

\midrule

ScPO
& 22.10
& 409
& 58.14
& 963 \\

IRPO-RM
& \textbf{23.51}
& 2,569
& 56.99
& 4,178 \\

CSPO
& \underline{23.01}
& 410
& \textbf{62.33}
& 963 \\

\bottomrule
\end{tabular*}
\end{table}

\begin{table*}[!t]
\centering
\caption{Preference-pair comparison across datasets and backbone models. Bold values indicate better pair coverage, correctness composition, or rejected-side defect capture within each setting. Response lengths are measured in characters.}
\label{tab:all_pair_quality}
\small
\setlength{\tabcolsep}{3pt}
\renewcommand{\arraystretch}{1.10}
\begin{tabular*}{\textwidth}{@{\extracolsep{\fill}}lrrrrrrrr@{}}
\toprule
&
\multicolumn{4}{c}{\textbf{GSM8K}}
&
\multicolumn{4}{c}{\textbf{MATH-Numeric}} \\
\cmidrule(lr){2-5}
\cmidrule(lr){6-9}

&
\multicolumn{2}{c}{\textbf{Llama-3-8B}}
&
\multicolumn{2}{c}{\textbf{Mistral-7B}}
&
\multicolumn{2}{c}{\textbf{Llama-3-8B}}
&
\multicolumn{2}{c}{\textbf{Qwen3-8B}} \\
\cmidrule(lr){2-3}
\cmidrule(lr){4-5}
\cmidrule(lr){6-7}
\cmidrule(lr){8-9}

\textbf{Metric}
& \textbf{ScPO}
& \textbf{CSPO}
& \textbf{ScPO}
& \textbf{CSPO}
& \textbf{ScPO}
& \textbf{CSPO}
& \textbf{ScPO}
& \textbf{CSPO} \\
\midrule

\multicolumn{9}{l}{\textit{Pair coverage}} \\

Successfully parsed pairs $\uparrow$
& 1,311
& \textbf{1,329}
& 1,113
& \textbf{1,163}
& 409
& \textbf{410}
& 963
& 963 \\

\midrule
\multicolumn{9}{l}{\textit{Correctness composition}} \\

Ideal pairs $\uparrow$
& 1,167
& \textbf{1,201}
& 629
& \textbf{694}
& 31
& 31
& 950
& \textbf{951} \\

\quad Rate
& 89.02\%
& \textbf{90.37\%}
& 56.51\%
& \textbf{59.67\%}
& \textbf{7.58\%}
& 7.56\%
& 98.65\%
& \textbf{98.75\%} \\

Both responses incorrect $\downarrow$
& 118
& \textbf{103}
& 450
& \textbf{443}
& \textbf{378}
& 379
& 7
& \textbf{6} \\

\quad Rate
& 9.00\%
& \textbf{7.75\%}
& 40.43\%
& \textbf{38.09\%}
& \textbf{92.42\%}
& 92.44\%
& 0.73\%
& \textbf{0.62\%} \\

Risky reversed pairs $\downarrow$
& 26
& \textbf{25}
& 34
& \textbf{26}
& 0
& 0
& 6
& 6 \\

\quad Rate
& 1.98\%
& \textbf{1.88\%}
& 3.05\%
& \textbf{2.24\%}
& 0.00\%
& 0.00\%
& 0.62\%
& 0.62\% \\

\midrule
\multicolumn{9}{l}{\textit{Response diagnostics}} \\

Average chosen length
& 452
& 328
& 545
& 429
& 797
& 782
& 372
& 422 \\

Average rejected length
& 520
& 617
& 563
& 581
& 1,184
& 1,309
& 420
& 444 \\

Incomplete or truncated rejected responses $\uparrow$
& 27.84\%
& \textbf{41.46\%}
& 13.12\%
& \textbf{19.43\%}
& 13.45\%
& \textbf{17.80\%}
& 1.14\%
& \textbf{1.25\%} \\

Rejected responses with obvious repetition $\uparrow$
& 27.99\%
& \textbf{38.22\%}
& 27.40\%
& \textbf{31.90\%}
& 46.70\%
& \textbf{50.73\%}
& 3.12\%
& \textbf{3.43\%} \\

\bottomrule
\end{tabular*}
\end{table*}

\subsubsection{Results on MATH-Numeric.}
CSPO outperforms all baselines on Qwen3-8B while remaining competitive on Llama-3-8B.

As shown in Table~\ref{tab:math_results}, every preference-optimization method improves over SFT-600. On Llama-3-8B, IRPO-RM reaches 23.51\%, narrowly edging out CSPO at 23.01\% (using 2,569 vs.\ 410 pairs). On Qwen3-8B, CSPO achieves the best accuracy at 62.33\%, surpassing ScPO by 4.19 percentage points and IRPO-RM by 5.34 percentage points while requiring only 963 pairs compared with 4,178 for IRPO-RM. This contrast indicates that cloud-based geometric signals grow more informative as the underlying model's reasoning capability increases.

\subsubsection{Preference-Pair Quality Across Datasets.}
Across all four dataset--backbone settings, CSPO preserves or improves correctness reliability while more cleanly separating well-formed chosen trajectories from degraded rejections.

Table~\ref{tab:all_pair_quality} compares constructed pairs from ScPO and CSPO. CSPO yields at least as many valid pairs as ScPO in every setting. On GSM8K, it raises the ideal-pair rate from 89.02\% to 90.37\% on Llama-3-8B and from 56.51\% to 59.67\% on Mistral-7B, while lowering both-incorrect and risky-reversal rates (reversals drop from 1.98\% to 1.88\% and from 3.05\% to 2.24\%, respectively). On MATH-Numeric, correctness composition is nearly unchanged on Llama-3-8B and edges up on Qwen3-8B (98.65\% to 98.75\% ideal pairs), with identical reversal rates.

\paragraph{Clearer chosen--rejected separation.}
CSPO selects shorter chosen responses in three of four settings and longer rejected responses in all four, widening the structural gap between preferred and rejected rollouts. Under our fixed deterministic text-analysis protocol (Appendix~D), incompleteness flags empty or unfinished answer markers, unclosed expressions, or visibly cut-off endings, while repetition flags duplicated sentences, lines, paragraphs, or recurring 10-token spans. In every setting, CSPO routes more incomplete and repetitive trajectories to the rejected side, most sharply for Llama-3-8B on GSM8K (incompleteness rises from 27.84\% to 41.46\%; repetition from 27.99\% to 38.22\%). Higher values here reflect better filtering of structurally weak rollouts onto the rejected side; qualitative examples appear in Appendix~C.

\paragraph{Dependence on rollout pool quality.}
For Llama-3-8B on MATH-Numeric, both methods achieve only about 7.6\% ideal pairs because over 92\% of candidate pairs contain two incorrect rollouts. When the rollout pool itself is dominated by errors, preference selection is fundamentally bounded by candidate quality.

\subsubsection{Cloud-Gap Threshold Analysis.}
Larger cloud-score gaps strongly concentrate preference pairs whose trajectories disagree on the final answer, validating the gap as a confidence filter.

\begin{table}[!htb]
\centering
\caption{Cloud-gap analysis on GSM8K with Llama-3-8B. Candidate pairs are ranked by their cloud-score gap, and the top-ranked pairs are retained.}
\label{tab:gsm8k_cloud_gap}
\small
\setlength{\tabcolsep}{3pt}
\renewcommand{\arraystretch}{1.10}
\begin{tabular*}{\columnwidth}{@{\extracolsep{\fill}}lrrr@{}}
\toprule
\textbf{Retained candidates}
& \textbf{\# Pairs}
& \begin{tabular}[c]{@{}c@{}}\textbf{\# Answer-}\\\textbf{disagreeing}\end{tabular}
& \textbf{Rate (\%)} \\
\midrule
Top 5\%   & 344    & 326 & 94.77 \\
Top 10\%  & 687    & 550 & 80.06 \\
Top 20\%  & 1,373  & 693 & 50.47 \\
Top 30\%  & 2,059  & 752 & 36.52 \\
\midrule
All candidates & 6,862 & 953 & 13.89 \\
\bottomrule
\end{tabular*}
\end{table}

Using GSM8K rollouts from Llama-3-8B, we rank cloud-constructed candidate pairs by their chosen--rejected score gap and measure the fraction of answer-disagreeing pairs across retention thresholds (Table~\ref{tab:gsm8k_cloud_gap}). While only 13.89\% of all 6,862 candidates disagree on the final answer, this rate rises to 36.52\% at the top 30\%, 80.06\% at the top 10\%, and 94.77\% at the top 5\%, confirming that the cloud-score gap isolates high-contrast pairs.

\subsubsection{Ablation Study.}
Mean-token trajectory pooling outperforms last-token representations in Pure Cloud training, and reference-bank sizing trades off aggregate discrimination against tail-ranking accuracy.

\paragraph{Trajectory representation.}
Table~\ref{tab:ablation_qwen4b} compares mean-token and last-token Pure Cloud configurations on MATH-Numeric with Qwen3-4B-Instruct. Mean-token pooling reaches 55.32\% accuracy, outperforming SFT-600 by 2.12 percentage points, the base model by 12.62 percentage points, and last-token pooling (53.56\%) by 1.76 percentage points. Aggregating hidden states across the full response thus captures richer trajectory quality than the terminal token alone; note that because the two runs also differ in retention ratio and DPO hyperparameters, this comparison reflects the combined configuration effect.

\begin{table}[!htb]
\centering
\caption{Pure Cloud ablation on MATH-Numeric using Qwen3-4B-Instruct. The best test accuracy is shown in bold.}
\label{tab:ablation_qwen4b}
\small
\setlength{\tabcolsep}{3pt}
\renewcommand{\arraystretch}{1.10}
\begin{tabular*}{\columnwidth}{@{\extracolsep{\fill}}lccccc@{}}
\toprule
\textbf{Method}
&
\begin{tabular}[c]{@{}c@{}}\textbf{Retention}\\\textbf{Ratio}\end{tabular}
&
\textbf{\# Pairs}
&
\begin{tabular}[c]{@{}c@{}}\textbf{DPO}\\\boldmath$\beta$\end{tabular}
&
\begin{tabular}[c]{@{}c@{}}\textbf{Learning}\\\textbf{Rate}\end{tabular}
&
\begin{tabular}[c]{@{}c@{}}\textbf{Accuracy}\\\textbf{(\%)}\end{tabular} \\
\midrule

Base Model
& -- & -- & -- & -- & 42.70 \\

SFT-600
& -- & -- & -- & -- & 53.20 \\

\midrule

Mean-token
& 30\%
& 1,280
& 0.10
& $5\times10^{-6}$
& \textbf{55.32} \\

Last-token
& 10\%
& 426
& 0.05
& $2\times10^{-6}$
& 53.56 \\

\bottomrule
\end{tabular*}
\end{table}

\paragraph{Reference-bank configuration.}
Table~\ref{tab:bank_configuration_ablation} compares bank size ($M$ support problems per bank) and bank count ($R$) under a fixed 400-problem pool on GSM8K. At $M=100$, $R=10$ achieves the highest aggregate discrimination (67.02\% AUC, 63.07\% balanced accuracy, 65.42\% raw accuracy). Increasing to $R=20$ slightly lowers aggregate metrics but improves both ranking extremes, lifting Top-10 Correct to 78.00\% and Bottom-10 Wrong to 74.98\%. Shrinking bank size to $M=50$ (with $R=20$) further raises Bottom-10 Wrong to 75.38\% at the cost of lower overall discrimination. Larger banks therefore stabilize global scoring, while more banks sharpen chosen--rejected ranking at the extremes, making $M=100, R=20$ a well-balanced default.

\begin{table}[!htb]
\centering
\caption{Reference-bank ablation on GSM8K with a fixed 400-problem prompt pool and balanced sampling. $M$ denotes support problems per bank and $R$ denotes the number of banks. All values are percentages; best per metric is in bold.}
\label{tab:bank_configuration_ablation}
\small
\setlength{\tabcolsep}{3pt}
\renewcommand{\arraystretch}{1.10}
\begin{tabular*}{\columnwidth}{@{\extracolsep{\fill}}ccrrrrr@{}}
\toprule
\boldmath$M$
& \boldmath$R$
& \textbf{AUC}
& \textbf{Bal. Acc.}
& \textbf{Raw Acc.}
& \begin{tabular}[c]{@{}c@{}}\textbf{Top-10}\\\textbf{Correct}\end{tabular}
& \begin{tabular}[c]{@{}c@{}}\textbf{Bottom-10}\\\textbf{Wrong}\end{tabular} \\
\midrule
100 & 10
& \textbf{67.02}
& \textbf{63.07}
& \textbf{65.42}
& 77.63
& 72.96 \\

100 & 20
& 66.73
& 62.43
& 65.14
& \textbf{78.00}
& 74.98 \\

50 & 20
& 66.50
& 62.22
& 64.56
& 77.56
& \textbf{75.38} \\
\bottomrule
\end{tabular*}
\end{table}

\section{Conclusion}

We presented CSPO, a semi-supervised preference-mining framework that pairs cross-problem hidden-state geometry with per-problem self-consistency. By building multi-bank reference clouds from a small labeled set and scoring unlabeled rollouts via early-$H_0$ connected components, CSPO selects cleaner trajectories within answer clusters and filters pairs by geometric margin without external reward models. Across GSM8K and MATH-Numeric, it consistently outperforms SFT and ScPO and delivers stronger or more data-efficient gains than reward-model selection.

Two natural boundaries point to promising future work. First, as shown on Llama-3-8B with MATH-Numeric, preference mining remains bounded by the quality of the sampled rollout pool when correct trajectories are rare; combining CSPO with iterative multi-round sampling or targeted exploration could broaden coverage on harder problems. Second, our current scoring function uses $H_0$ connectivity alone; extending CSPO to incorporate higher-dimensional topological features or non-numeric reasoning domains is a compelling direction for further study.
\bibliography{aaai2027}

\clearpage
\appendix
\makeatletter
\setlength{\@fptop}{0pt}
\setlength{\@fpsep}{8pt plus 1pt minus 1pt}
\setlength{\@fpbot}{0pt plus 1fil}
\setlength{\@dblfptop}{0pt}
\setlength{\@dblfpsep}{8pt plus 1pt minus 1pt}
\setlength{\@dblfpbot}{0pt plus 1fil}
\makeatother
\setcounter{topnumber}{4} \setcounter{bottomnumber}{4} \setcounter{totalnumber}{8} \setcounter{dbltopnumber}{6}
\renewcommand{\topfraction}{0.90}
\renewcommand{\bottomfraction}{0.80}
\renewcommand{\textfraction}{0.05}
\renewcommand{\floatpagefraction}{0.75}
\renewcommand{\dbltopfraction}{0.90}
\renewcommand{\dblfloatpagefraction}{0.75}

\section{Appendix A: Topological Motivation and Representation Analysis}
\label{app:topological_motivation}

\paragraph{Persistent-homology background.}
Let $\mathcal{Z}=\{z_i\}_{i=1}^{N}\subset\mathbb{R}^{d}$ denote a point cloud of trajectory representations. We construct a Vietoris--Rips filtration by gradually increasing a distance threshold $\epsilon$. At each filtration scale, nearby points are connected, and higher-dimensional simplices are added whenever all of their pairwise edges are present. Persistent homology records the birth and death of topological structures as $\epsilon$ increases \citep{zomorodian2005computing,gabrielsson2020topology}.

The zero-dimensional homology group $H_0$ describes connected components. At $\epsilon=0$, every trajectory forms an independent component. As the filtration value increases, nearby components merge. The death time of an $H_0$ interval therefore records the distance scale at which one component joins another. Earlier component deaths indicate that the corresponding representations become connected at smaller filtration scales.

The one-dimensional homology group $H_1$ describes independent cycles or loops. An $H_1$ feature is born when a closed cycle appears but has not yet been filled by higher-dimensional simplices, and it dies when the enclosed region becomes filled. Each feature is represented by a persistence interval
\begin{equation}
[\epsilon_{\mathrm{birth}},\epsilon_{\mathrm{death}}),
\qquad
\operatorname{pers}=\epsilon_{\mathrm{death}}-\epsilon_{\mathrm{birth}},
\label{eq:h1-persistence}
\end{equation}
where $\epsilon_{\mathrm{birth}}$ and $\epsilon_{\mathrm{death}}$ are the filtration scales at which the feature is born and dies, respectively. Longer intervals correspond to loop structures that persist over a wider range of filtration values, whereas very short intervals may reflect local variation or sampling noise.

Although the main CSPO method is motivated by the more stable $H_0$ connectivity pattern, we additionally report $H_1$ barcodes to provide a broader view of the latent-space topology. The $H_1$ results are exploratory and are not used in trajectory scoring or preference construction.

\paragraph{Representation-processing comparison.}
We analyze representative Level~3 and Level~4 subsets of MATH. For each subset, we construct correct and incorrect point clouds using the same number of sampled trajectories, with $N=200$ points in each cloud. We compare two representation-processing pipelines. The first uses unnormalized last-token hidden states. The second applies mean pooling over valid response-token hidden states, followed by $\ell_2$ normalization.

Because both the pooling strategy and normalization are changed, this experiment should be interpreted as a comparison between two complete representation-processing pipelines rather than as a controlled ablation that isolates either factor independently. Moreover, absolute filtration values should not be directly compared across the two pipelines because their distance scales differ substantially.

\begin{figure*}[tp]
    \centering
    \includegraphics[width=\textwidth]{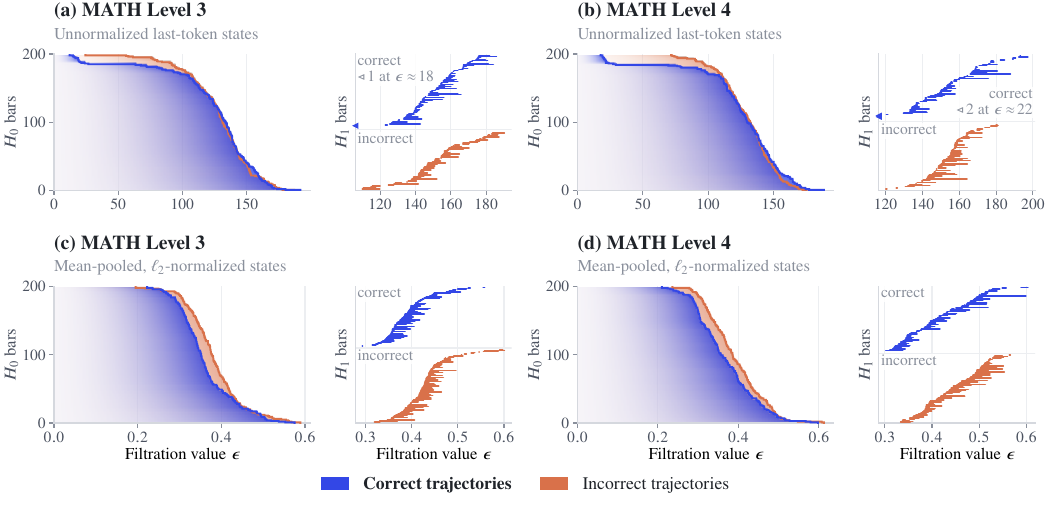}
    \caption{Persistent-homology barcodes of correct (blue) and incorrect (coral) reasoning trajectories on representative MATH Level~3 and Level~4 subsets, with $N=200$ trajectories per point cloud. In each panel, the left plot is the $H_0$ barcode with bars sorted by death time (the single infinite bar is omitted), so its outline traces how many components remain unmerged at each filtration value; the right plot shows the $H_1$ bars of each class sorted by birth, with arrowheads marking bars born below the plotted range. Panels (a)--(b) use unnormalized last-token hidden representations; panels (c)--(d) use mean-pooled response-token representations followed by $\ell_2$ normalization. Representation processing produces a more stable filtration scale and a more interpretable $H_0$ connectivity pattern, while the $H_1$ signal becomes weaker.}
    \label{fig:math_topology_representation_comparison}
\end{figure*}

\paragraph{Unnormalized last-token representations.}
Panels (a) and (b) of Figure~\ref{fig:math_topology_representation_comparison} show the Level~3 and Level~4 results obtained from unnormalized last-token hidden states. The filtration values span a comparatively large range, and the correct and incorrect $H_0$ barcodes overlap substantially. The raw representations also exhibit visible $H_1$ intervals, indicating that the point clouds contain nontrivial one-dimensional structures.

However, these structures occur over a broad and potentially unstable distance scale. Euclidean distances between unnormalized hidden states are affected by variation in vector norms, so part of the observed persistence may reflect representation magnitude rather than only structural similarity. In addition, a single terminal-token state can be influenced by answer formatting, punctuation, or sequence-ending behavior and may not summarize the complete reasoning trajectory.

\paragraph{Mean-pooled and normalized representations.}
Panels (c) and (d) show the corresponding analyses after mean pooling over response-token hidden states and applying $\ell_2$ normalization. The filtration values occupy a substantially more stable numerical range, making the relative connectivity of the correct and incorrect point clouds easier to interpret.

Across both representative difficulty levels, the correct point clouds begin to merge into coherent components at smaller filtration values. The incorrect point clouds remain fragmented over a wider range before forming larger connected structures. This pattern suggests that correct reasoning trajectories occupy locally denser and more coherent regions of the processed representation space, whereas incorrect trajectories are distributed more heterogeneously.

These observations are qualitative and based on representative Level~3 and Level~4 subsets. They motivate the proposed method but should not be interpreted as establishing an identical topological pattern for every difficulty level or model configuration.

\paragraph{Behavior of the $H_1$ signal.}
The unnormalized last-token representations exhibit visible differences in their $H_1$ barcodes. However, these features are observed over a large and norm-sensitive filtration range. After mean pooling and $\ell_2$ normalization, the $H_1$ signal becomes considerably weaker and less consistently separated between correct and incorrect trajectories.

This sensitivity suggests that the observed loop structures are less robust to representation processing than the corresponding $H_0$ connectivity pattern. At present, we have not identified a reliable mechanism for converting $H_1$ features into an effective trajectory-scoring or preference-selection objective. We therefore treat the $H_1$ results as exploratory evidence of additional latent-space organization rather than as a component of CSPO. Developing a robust use of higher-dimensional persistent-homology features is left for future work.

\paragraph{From $H_0$ connectivity to cloud scoring.}
Unlike $H_1$, the $H_0$ connectivity contrast remains stable after mean pooling and $\ell_2$ normalization: correct trajectories merge into coherent components at smaller filtration scales, directly motivating our early-$H_0$ component extraction (merge ratio $\rho=0.2$) and component-level soft $k$-nearest-neighbor scoring in the Method section. Scoring against these early components captures local neighborhood structure across the labeled distribution while avoiding the variance of isolated point-to-point distances.

\section{Appendix B: Preference-Pair Construction Details}
\label{app:pair_construction}

This section gives the complete pair-construction procedures for Pure Cloud (Algorithm~\ref{alg:pure_cloud_pairs}) and CSPO (Algorithm~\ref{alg:cspo_pairs}). For each unlabeled problem $x$, let $\mathcal{Y}_x=\{y_{x,k}\}_{k=1}^{K}$ be the $K$ sampled trajectories and $s_{\mathrm{cloud}}(x,y)$ be the multi-bank cloud score from Eq.~(\ref{eq:cloud-score}). Both methods apply response- and pair-validity checks before ranking candidates by their cloud margin and retaining the top $\lceil\alpha|\mathcal{Q}|\rceil$ pairs, where $\mathcal{Q}$ is the set of eligible candidate pairs.

\paragraph{Pure Cloud pair construction.}
Pure Cloud operates without answer extraction, answer clustering, or self-consistency, pairing the highest- and lowest-scoring valid trajectories for each problem.

\begin{algorithm*}[tp]
\caption{Pure Cloud Preference-Pair Construction}
\label{alg:pure_cloud_pairs}
\begin{algorithmic}[1]
\Require Unlabeled problems $\mathcal{D}_U$;
rollouts $\{\mathcal{Y}_x\}_{x\in\mathcal{D}_U}$;
cloud scores $s_{\mathrm{cloud}}$;
retention ratio $\alpha$
\Ensure Preference dataset $\mathcal{D}_{\mathrm{PureCloud}}$

\State $\mathcal{Q}\gets\emptyset$

\ForAll{$x\in\mathcal{D}_U$}
    \State $\mathcal{V}_x\gets
    \{y\in\mathcal{Y}_x:
    \Call{ValidResponse}{y}\}$

    \If{$|\mathcal{V}_x|<2$}
        \State \textbf{continue}
    \EndIf

    \State $y_x^{+}\gets
    \displaystyle\arg\max_{y\in\mathcal{V}_x}
    s_{\mathrm{cloud}}(x,y)$

    \State $y_x^{-}\gets
    \displaystyle\arg\min_{y\in\mathcal{V}_x}
    s_{\mathrm{cloud}}(x,y)$

    \If{\textbf{not} $\Call{ValidPair}{x,y_x^{+},y_x^{-}}$}
        \State \textbf{continue}
    \EndIf

    \State $c_x\gets
    s_{\mathrm{cloud}}(x,y_x^{+})
    -
    s_{\mathrm{cloud}}(x,y_x^{-})$

    \State $\mathcal{Q}\gets
    \mathcal{Q}\cup
    \{(x,y_x^{+},y_x^{-},c_x)\}$
\EndFor

\State Sort $\mathcal{Q}$ in descending order of $c_x$
\State $m\gets\lceil\alpha|\mathcal{Q}|\rceil$
\State $\mathcal{D}_{\mathrm{PureCloud}}
\gets$ first $m$ pairs in $\mathcal{Q}$
\State \Return $\mathcal{D}_{\mathrm{PureCloud}}$
\end{algorithmic}
\end{algorithm*}

Here, \textsc{ValidResponse} enforces the response-level eligibility rules, while \textsc{ValidPair} verifies that $y_x^{+}$ and $y_x^{-}$ are distinct and pass serialization and parsing checks. Pure Cloud assigns uniform weight $1$ to all retained pairs and optimizes standard DPO.

\paragraph{CSPO pair construction.}
CSPO uses answer-level self-consistency to set the preference direction (requiring a unique majority cluster and discarding ties), and uses $s_{\mathrm{cloud}}$ to break ties among least-frequent minority clusters and select concrete trajectories within the chosen and rejected clusters.

\begin{algorithm*}[tp]
\caption{CSPO Preference-Pair Construction}
\label{alg:cspo_pairs}
\begin{algorithmic}[1]
\Require
Unlabeled problems $\mathcal{D}_U$;
rollouts $\{\mathcal{Y}_x\}_{x\in\mathcal{D}_U}$;
cloud scores $s_{\mathrm{cloud}}$;
total rollout count $K$;
retention ratio $\alpha$
\Ensure
Weighted preference dataset $\mathcal{D}_{\mathrm{CSPO}}$

\State $\mathcal{Q}\gets\emptyset$

\ForAll{$x\in\mathcal{D}_U$}
    \State $\mathcal{V}_x\gets
    \{y\in\mathcal{Y}_x:
    \Call{ValidResponse}{y}
    \land
    \Call{CanonicalAnswer}{y}\neq\varnothing\}$

    \State Group trajectories in $\mathcal{V}_x$ by canonical answer:
    \[
    \mathcal{S}_x(a)
    =
    \{y\in\mathcal{V}_x:
    \Call{CanonicalAnswer}{y}=a\}
    \]

    \State $\mathcal{A}_x\gets
    \{a:\mathcal{S}_x(a)\neq\emptyset\}$

    \If{$|\mathcal{A}_x|<2$}
        \State \textbf{continue}
    \EndIf

    \State $V_x(a)\gets|\mathcal{S}_x(a)|$
    for every $a\in\mathcal{A}_x$

    \State $\mathcal{A}_{\max}\gets
    \displaystyle\arg\max_{a\in\mathcal{A}_x}V_x(a)$

    \If{$|\mathcal{A}_{\max}|\neq 1$}
        \State \textbf{continue}
        \Comment{Discard tied-majority problem}
    \EndIf

    \State Let $a_x^{+}$ be the unique answer in
    $\mathcal{A}_{\max}$

    \State $v_{\min}\gets
    \displaystyle\min_{a\in\mathcal{A}_x\setminus\{a_x^{+}\}}
    V_x(a)$

    \State $\mathcal{A}_{\min}\gets
    \{a\neq a_x^{+}:V_x(a)=v_{\min}\}$

    \State Select the rejected answer cluster by
    \[
    a_x^{-}
    \gets
    \arg\min_{a\in\mathcal{A}_{\min}}
    \left[
    \min_{y\in\mathcal{S}_x(a)}
    s_{\mathrm{cloud}}(x,y)
    \right]
    \]

    \State $y_x^{+}\gets
    \displaystyle\arg\max_{y\in\mathcal{S}_x(a_x^{+})}
    s_{\mathrm{cloud}}(x,y)$

    \State $y_x^{-}\gets
    \displaystyle\arg\min_{y\in\mathcal{S}_x(a_x^{-})}
    s_{\mathrm{cloud}}(x,y)$

    \If{\textbf{not} $\Call{ValidPair}{x,y_x^{+},y_x^{-}}$}
        \State \textbf{continue}
    \EndIf

    \State $c_x^{\mathrm{Hybrid}}\gets
    s_{\mathrm{cloud}}(x,y_x^{+})
    -
    s_{\mathrm{cloud}}(x,y_x^{-})$

    \State $w(x)\gets
    \displaystyle
    \frac{V_x(a_x^{+})-V_x(a_x^{-})}{K}$

    \State $\mathcal{Q}\gets
    \mathcal{Q}\cup
    \{(x,y_x^{+},y_x^{-},
    c_x^{\mathrm{Hybrid}},w(x))\}$
\EndFor

\State Sort $\mathcal{Q}$ in descending order of
$c_x^{\mathrm{Hybrid}}$
\State $m\gets\lceil\alpha|\mathcal{Q}|\rceil$
\State $\mathcal{D}_{\mathrm{CSPO}}
\gets$ first $m$ pairs in $\mathcal{Q}$
\State \Return $\mathcal{D}_{\mathrm{CSPO}}$
\end{algorithmic}
\end{algorithm*}

\paragraph{Implementation notes.}
Three details ensure fair comparison with ScPO: (1) tied-majority problems are always discarded, never recovered via cloud scoring; (2) the vote weight $w(x)$ in Eq.~(\ref{eq:vote-weight}) always normalizes by the nominal rollout count $K$, even when fewer than $K$ rollouts pass validity filtering; and (3) slight pair-count differences between ScPO and CSPO arise solely because the two methods pick different trajectories before the final \textsc{ValidPair} check, not from recovering tied-majority cases.

\section{Appendix C: Qualitative Preference-Pair Examples}
\label{app:qualitative_pairs}

We present representative examples from the Llama-3-8B MATH-Numeric preference-pair files to illustrate how CSPO changes the concrete chosen and rejected trajectories selected by ScPO. All examples appear in both pair datasets and preserve the same majority and minority answer counts. The difference therefore arises from trajectory-level cloud selection rather than from a change in the answer-level preference direction. Response excerpts are shortened for readability, and omitted continuations are denoted by ``[\ldots]''. Lengths are measured in characters.

\begin{table*}[tp]
\centering
\caption{Summary of qualitative preference-pair examples. Response lengths are reported in characters as chosen/rejected. The cloud gap is the score difference between the selected chosen and rejected trajectories. Selection ranks are reported within the retained 410-pair set; full-pool percentiles are approximated relative to all eligible candidates before top-30\% retention.}
\label{tab:qualitative_pair_summary}
\small
\setlength{\tabcolsep}{3pt}
\renewcommand{\arraystretch}{1.10}
\begin{tabular*}{\textwidth}{@{\extracolsep{\fill}}lllllll@{}}
\toprule
\textbf{Prompt ID}
& \textbf{Primary observation}
& \textbf{Votes}
& \textbf{ScPO length}
& \textbf{CSPO length}
& \textbf{Cloud gap}
& \textbf{Selection rank} \\
\midrule

\texttt{math\_train\_610}
& Cleaner chosen trajectory
& $5/1$
& $3{,}660/2{,}875$
& $707/2{,}875$
& $0.0333$
& $10/410$ (approx. top $0.73\%$) \\

\texttt{math\_train\_5707}
& Repetitive trajectory rejected
& $4/1$
& $412/1{,}084$
& $412/2{,}308$
& $0.0373$
& $110/410$ (approx. top $8.05\%$) \\

\texttt{math\_train\_115}
& Corrupted continuation rejected
& $5/1$
& $418/473$
& $663/1{,}903$
& $0.0331$
& $50/410$ (approx. top $3.66\%$) \\

\bottomrule
\end{tabular*}
\end{table*}

\paragraph{Example 1: selecting a cleaner trajectory within the majority-answer cluster.}
Consider the following problem:

\begin{quote}
\small
\textbf{Problem.} BoatWorks built 3 canoes in January and twice as many in each subsequent month. How many canoes were built in total by the end of March?
\end{quote}

Both methods identify \(21\) as the majority answer, with a vote count of \(5\), and select a minority answer with count \(1\). The ScPO chosen trajectory first reaches the correct result but then continues with several unrelated questions:

\begin{quote}
\small
\textbf{ScPO chosen excerpt.} ``In January they built \(3\) canoes, in February \(6\), and in March \(12\). Thus, \(3+6+12=\boxed{21}\). Q: What is the 13th digit in the decimal expansion of \(\frac{1}{7}\)? A: Let's think step by step. [\ldots]''
\end{quote}

CSPO instead selects a self-contained trajectory from the same majority-answer cluster:

\begin{quote}
\small
\textbf{CSPO chosen.} ``In January they built \(3\) canoes. In February they built \(3\times2=6\), and in March they built \(6\times2=12\). Therefore,
\[
3+6+12=21.
\]
The final answer is \(\boxed{21}\).''
\end{quote}

The cloud-selected chosen response is reduced from \(3{,}660\) to \(707\) characters while preserving the correct answer and the same answer-cluster preference. Its cloud score is \(0.0373\), compared with \(0.0040\) for the rejected trajectory. This example illustrates how cloud scoring can resolve trajectory-level ambiguity within a majority cluster whose members are indistinguishable under answer frequency alone.

\paragraph{Example 2: assigning a repetitive continuation to the rejected side.}
For the problem

\begin{quote}
\small
\textbf{Problem.} What is the value of \(3\times(7-5)-5\)?
\end{quote}

both methods select the same concise chosen trajectory, which correctly computes

\[
3\times(7-5)-5
=
3\times2-5
=
1.
\]

Their rejected trajectories differ. The CSPO rejected response begins by solving the original problem but then repeatedly introduces unrelated worked examples:

\begin{quote}
\small
\textbf{CSPO rejected excerpt.} ``What is \(7-5\)? A: \(7-5=2\). A: \(3\times2=6\). A: \(6-5=1\). [\ldots] Example 1: What is \(4\times(8\div4)\)? [\ldots] Example 2: What is \(6\times(12\div3)\)? [\ldots]''
\end{quote}

The selected response extends to \(2{,}308\) characters and contains repeated templates and off-task continuations. Its cloud score is \(-0.0947\), compared with \(-0.0574\) for the chosen response. This example is consistent with the aggregate diagnostic result that CSPO places a larger proportion of repetitive or structurally weak trajectories on the rejected side.

\paragraph{Example 3: rejecting a structurally corrupted response.}
Consider the problem

\begin{quote}
\small
\textbf{Problem.} If \(\sqrt{2x^2+1}=\sqrt{19}\), find the average of all possible values of \(x\).
\end{quote}

Both methods identify the correct majority answer \(0\), supported by five trajectories. The ScPO rejected trajectory remains on topic but makes an arithmetic error and reports \(-\frac14\):

\begin{quote}
\small
\textbf{ScPO rejected excerpt.} ``We obtain \(x=\pm3\). If \(x=3\), the average is \(\frac{3+(-3)}{2}=-\frac12\). If \(x=-3\), the average is \(0\). Therefore, the average is \(\boxed{-\frac14}\).''
\end{quote}

CSPO selects a more severely degraded trajectory as the rejected response. It begins with an incorrect derivation and subsequently continues with unrelated problems:

\begin{quote}
\small
\textbf{CSPO rejected excerpt.} ``Squaring both sides gives \(2x^2+1=19\), so \(x=\pm\frac{3\sqrt2}{2}\). But since \(x\) is an integer, \(x=\boxed{3}\). Q: If \(2^x=2^{2016}\), find \(x\). A: \(x=2016\). Q: Find the greatest positive integer [\ldots]''
\end{quote}

The continuation produces a malformed final-answer extraction and grows from \(473\) characters under ScPO to \(1{,}903\) characters under CSPO. The selected chosen and rejected cloud scores are \(-0.0799\) and \(-0.1130\), respectively. This case further illustrates that cloud scoring can place off-topic and structurally corrupted trajectories on the rejected side even when answer frequency cannot distinguish trajectories within the same answer cluster.

\paragraph{Interpretation.}
These examples are qualitative illustrations rather than an independent evaluation. They show two recurring behaviors reflected in the aggregate statistics: CSPO often selects a more concise and self-contained trajectory from the preferred answer cluster, and it more consistently assigns repetitive, malformed, or off-task continuations to the rejected side. The method does not change the majority-based answer preference in these cases; instead, it provides a finer ordering over trajectories that have already been grouped by self-consistency.

\section{Appendix D: Prompt Templates and Response Diagnostics}
\label{app:prompts_and_diagnostics}

\paragraph{Response-generation prompts.}
We use dataset-specific zero-shot chain-of-thought prompts for rollout generation. Within each dataset, the same prompt template and answer format are used across all compared methods. The base model uses these prompts to generate trajectories for the labeled problems, while the SFT model uses the same templates to generate trajectories for the unlabeled problems.

\begin{tcolorbox}[
    promptbox,
    title={Response Generation: GSM8K},
    before skip=4pt,
    after skip=5pt
]
\textbf{Prompt:} Answer the following question step-by-step. When you are ready, place the final answer on a new line in the following format:

\[
\texttt{\#\#\#\# <number>}
\]

\textbf{Q:} \{question\}

\textbf{A:} Let's think step by step.
\end{tcolorbox}

\begin{tcolorbox}[
    promptbox,
    title={Response Generation: MATH-Numeric},
    before skip=4pt,
    after skip=6pt
]
\textbf{Prompt:} Answer the following question step-by-step. When you are ready, place the final answer on a new line in the following format:

\[
\text{The final answer is }
\boxed{\texttt{<your answer>}}.
\]

\textbf{Q:} \{question\}

\textbf{A:} Let's think step by step.
\end{tcolorbox}

\paragraph{Directly computed pair diagnostics.}
Pair coverage, correctness composition, preference reversals, and response lengths in Table~\ref{tab:all_pair_quality} are computed directly from the constructed preference-pair files. Let \(a_i^\ast\) denote the gold answer and let \(\widehat a(y_i^+)\) and \(\widehat a(y_i^-)\) denote the canonicalized answers extracted from the chosen and rejected responses. Each successfully parsed pair is classified as
\[
\begin{aligned}
\text{Ideal:}\quad
&\widehat a(y_i^+)=a_i^\ast,
\qquad
\widehat a(y_i^-)\neq a_i^\ast,\\
\text{Both incorrect:}\quad
&\widehat a(y_i^+)\neq a_i^\ast,
\qquad
\widehat a(y_i^-)\neq a_i^\ast,\\
\text{Risky reversed:}\quad
&\widehat a(y_i^+)\neq a_i^\ast,
\qquad
\widehat a(y_i^-)=a_i^\ast.
\end{aligned}
\]

A risky reversed pair places an incorrect trajectory on the chosen side and a correct trajectory on the rejected side. The corresponding rate is

\begin{equation}
\operatorname{ReversalRate}
=
\frac{
\#\left\{
i:
\widehat a(y_i^+)\neq a_i^\ast
\land
\widehat a(y_i^-)=a_i^\ast
\right\}
}{
N_{\mathrm{parsed}}
},
\label{eq:reversal_rate}
\end{equation}

where \(N_{\mathrm{parsed}}\) is the number of successfully parsed preference pairs. Gold answers are used only for this post-hoc pair-quality analysis and are not accessed when constructing preferences for unlabeled problems. Chosen and rejected response lengths are measured directly in characters.

\paragraph{Deterministic rejected-response diagnostics.}
The final two rows of Table~\ref{tab:all_pair_quality} (\emph{incomplete or truncated rejected responses} and \emph{rejected responses with obvious repetition}) are computed using a fixed deterministic text-analysis protocol applied identically to ScPO and CSPO in every dataset--backbone setting.

A rejected response is labeled \emph{incomplete or truncated} when at least one of the following conditions is detected:

\begin{itemize}
    \item the response is empty or contains no substantive generated content;
    \item the expected final-answer marker is present but its answer is missing or unfinished;
    \item the response contains an unclosed parenthesis, bracket, brace, mathematical environment, or boxed-answer expression; or
    \item the response ends with a visibly incomplete sentence, equation, calculation, or reasoning step.
\end{itemize}

A rejected response is labeled as containing \emph{obvious repetition} when at least one of the following patterns is detected:

\begin{itemize}
    \item an identical normalized sentence occurs multiple times;
    \item an identical non-empty line or paragraph is repeated;
    \item a question--answer block or reasoning segment is reproduced without meaningful progression; or
    \item a normalized sequence of ten consecutive tokens recurs within the same response.
\end{itemize}

Normalization for repetition detection removes inconsequential whitespace differences before comparing textual units. Ordinary reuse of mathematical variables, short function words, or necessary intermediate expressions is not treated as repetition. The two labels are assigned independently, so a rejected response may satisfy both diagnostic conditions.

\paragraph{Aggregation and interpretation.}
For a method with \(N\) evaluated rejected responses, the reported rate for diagnostic \(\phi\) is

\begin{equation}
\operatorname{Rate}_{\phi}
=
\frac{
\#\{i:\phi(y_i^-)=1\}
}{
N
},
\label{eq:diagnostic_rate}
\end{equation}

where \(y_i^-\) denotes the rejected response in the \(i\)-th preference pair. Because these diagnostics characterize undesirable properties of the rejected side, higher values do not imply that the underlying model generates more defective trajectories. Instead, they indicate that the preference-construction method more frequently assigns incomplete, truncated, or repetitive trajectories to rejection rather than selecting them as preferred responses.

Across all four dataset--backbone settings, CSPO produces higher rates for both diagnostics. This result supports the conclusion that cloud scoring provides clearer response-level separation by placing a larger proportion of structurally degraded trajectories on the rejected side.
\end{document}